# Deep Adaptation of Adult-Child Facial Expressions by Fusing Landmark Features

Megan A. Witherow, *Member, IEEE,* Manar D. Samad, *Member, IEEE*, Norou Diawara, Haim Y. Bar, and Khan M. Iftekharuddin, *Senior Member, IEEE*

**Abstract**—Imaging of facial affects may be used to measure psychophysiological attributes of children through their adulthood for applications in education, healthcare, and entertainment, among others. Deep convolutional neural networks show promising results in classifying facial expressions of adults. However, classifier models trained with adult benchmark data are unsuitable for learning child expressions due to discrepancies in psychophysical development. Similarly, models trained with child data perform poorly in adult expression classification. We propose domain adaptation to concurrently align distributions of adult and child expressions in a shared latent space for robust classification of either domain. Furthermore, age variations in facial images are studied in age-invariant face recognition yet remain unleveraged in adult-child expression classification. We take inspiration from multiple fields and propose deep adaptive FACial Expressions fusing BEtaMix SElected Landmark Features (FACE-BE-SELF) for adult-child expression classification. For the first time in the literature, a mixture of Beta distributions is used to decompose and select facial features based on correlations with expression, domain, and identity factors. We evaluate FACE-BE-SELF using 5-fold cross validation for two pairs of adult-child data sets. Our proposed FACE-BE-SELF approach outperforms transfer learning and other baseline domain adaptation methods in aligning latent representations of adult and child expressions.

---

## 1 Introduction

From infancy to adulthood, facial expressions are a ubiquitous, information-rich component of human social interactions. Facial expressions may provide valuable information about a child's development into an adult [1], [2], [3]. Many applications of automatic facial expression analysis (FEA), including education (e.g., engagement in the classroom [4], [5], [6]), healthcare (e.g., monitoring of pain [7], [8], mental health [9], [10], autism [11], [12], [13]), and entertainment (e.g., video games [14], [15]) remain relevant from childhood into adulthood. To better support such applications, FEA models need to generalize across distinctive expression patterns from early childhood to adulthood. Developing models robust to age variations is a challenging problem in FEA [16], [17]. Most existing approaches optimize the FEA performance on data sets representing specific age ranges. There has been limited work on classifying facial expressions across age groups. Furthermore, age variations in facial images have been well-studied in facial age estimation and age-invariant face recognition (AIFR), but there has been little cross-pollination among these relevant research areas to improve FEA considering adult-child age variations. In the following sections, we discuss related work on the classification of adult and child expressions and methods from relevant research fields. Then, we propose a novel deep feature adaptation approach to the classification of adult and child expressions inspired by the state-of-the-art domain adaptation learning, facial age estimation, and AIFR literature.

### 1.1 Related work

#### 1.1.1 Classification of adult & child facial expressions

Existing off-the-shelf FEA tools and research [18], [19], [20] have been mostly developed using adult benchmark data sets [17], [21], [22]. However, facial morphology and kinematics gradually develop throughout childhood [23], [24], resulting in a distribution shift between child and adult expression patterns. For models trained on adult data sets, the distribution shift toward adults poorly generalizes distinctive patterns in child expressions [25], [26], [27], [28]. While benchmark data sets of child facial expressions remain limited, they are growing in number [29], [30], [31]. Therefore, there has been an emerging trend directed at the classification of child facial expressions [25], [26], [27], [28]. Recently, deep transfer learning using convolutional neural networks (CNNs) has shown promise for child facial expression classification [26], [27], [32]. However, recent

- *M.A. Witherow is with the Vision Lab, Department of Electrical & Computer Engineering, Old Dominion University, Norfolk, VA 23529. E-mail: mwith010@odu.edu.*
- *M.D. Samad is with the Department of Computer Science, Tennessee State University, Nashville, TN 37209. E-mail: msamad@tnstate.edu.*
- *N. Diawara is with the Department of Mathematics & Statistics, Old Dominion University, Norfolk, VA 23529. E-mail: ndiawara@odu.edu.*
- *H.Y. Bar is with the Department of Statistics, University of Connecticut, Storrs, CT 06269. E-mail: haim.bar@uconn.edu.*
- *K.M. Iftekharuddin is with the Vision Lab, Department of Electrical & Computer Engineering, Old Dominion University, Norfolk, VA 23529. E-mail: kiftekha@odu.edu.*

studies focus only on maximizing performance on child facial expression benchmarks, bounded by limited age range and sample size [17]. Such models tuned for child expressions fail to generalize to adult expressions [28]. To overcome the poor generalization problem across age groups, limited existing work on facial expression classification involving mixed age groups (child, adult, elderly) suggests two primary approaches: (1) curating a mixed age training set to match the age distribution of the test set [33], and (2) classifying images into age groups to determine the age-appropriate model for subsequent classification [34]. The first approach requires the age distribution of the test set to be known a priori with availability of benchmark data matching. The second approach requires a robust age group classifier to select an appropriate expression classifier model and benchmark data to train expression classifiers for individual age groups. Age group classification is a challenging problem [35], [36] and variations in expression make accurate age estimation even more challenging [34], [37]. Furthermore, developments in both facial structure and muscle movements contribute to visual differences in child and adult expressions. A child's growth is a gradual and uniquely individual process, making the transition unclear when a child manifests the full spectrum of adult expressions.

Recently, domain adaptation has shown an interesting pathway to adapt an adult expression classification model using few child expression samples [27]. This approach utilizes a dual stream deep CNN architecture and semantically aligns the class conditional distributions of child and adult domains [27]. The underlying framework of this approach [38] is based on learning a domain-invariant latent representation. Such domain-invariant representations have shown to generalize even to unseen domains [38]. We hypothesize that learning a domain-invariant representation of expressions may prove effective for facial expression classification across child and adult domains.

### 1.1.2 Recognition of age-varying facial images

While limited attention has been given to facial expression classification across age groups, facial age estimation [39] and AIFR [40] are active research areas. State-of-art approaches for facial age estimation and AIFR benefit from deep learning and fusion of geometric and texture features [36], [39], [41]. Geometric features derived from facial landmarks capture structural changes associated with childhood development while texture features capture skin artifacts, such as wrinkles, associated with adult aging [36], [41]. Contemporary studies continue to use traditional feature extraction methods, e.g., local binary patterns, histograms of oriented gradients, etc., but recently emphasize deep learning, e.g., CNNs, for texture feature extraction [39], [40], [42]. Common geometric landmark features include distances between landmarks, ratios of distances, and areas and angles of triangles formed by landmark triplets [35], [43], [44], [45]. Similar landmark features, including pairwise distances between landmarks and areas/angles of facial polygons formed by connecting neighboring landmark points, have also shown to be discriminative for FEA [46], [47], [48], [49]. Therefore, we hypothesize that domain-invariant representation learning of adult and child facial expressions can benefit from a fusion of CNN-extracted and landmark-derived features.

The use of the same feature types in both facial age estimation and AIFR suggests a subset of features correlated with and invariant to age. Statistical latent variable models optimized using the Expectation-Maximization (EM) algorithm have been applied to AIFR to decompose feature sets into age and identity factors [40]. This approach identifies a set of discriminative features for identity recognition using the identity factor, representing facial identity features invariant to age [40]. Gong et al. [50] have first proposed this approach using hidden factor analysis (HFA). HFA assumes the independence of age and identity, which is untrue in practice as different individuals may have different aging patterns [40]. To overcome the independence assumption, the modified HFA (MHFA) approach introduces an additional factor representing age and identity-correlated facial appearance variations [51]. Given that the appearance of facial expressions varies among individuals and age groups, we hypothesize that FEA can benefit from decomposition of feature sets into those correlated with expression, domain (adult or child), and identity. However, MHFA assumes that data are independent and identically distributed (i.i.d.) following a normal distribution with homogenous variances, which may not be true for real world facial expression data. Furthermore, HFA and MHFA require the optimization of one model per feature, making high dimensional feature vectors computationally prohibitive [50], [51]. Thus, principal component analysis (PCA) has been used for dimensionality reduction prior to HFA or MHFA [50], [51]. While PCA guarantees that the first principal components explain more of the variance than subsequent principal components, such linear data projection method does not guarantee that the PCA feature space will be discriminative for classification. Each principal component is a linear combination of all input features, making it less intuitive to understand the contribution of individual features. Moreover, all features, even those with limited contribution to discriminability, are needed to reproduce the same principal components.

Very recently, the Beta-Mixture (BetaMix) method [52] has been proposed to determine significant correlations among large numbers of variables using a mixture of Beta distributions. The method, based on ideas and results from convex geometry, works well even for moderate sample sizes, e.g., $N = 10$ depending on the number of predictors, and does not require assumptions of i.i.d., normality, or homogeneity of variances. The BetaMix method detects correlations among all the features at once, so the EM algorithm needs to be applied only once for all features rather than for individual features. Since the BetaMix method is appropriate for large feature vectors, dimensionality reduction is not required and the feature correlations may be interpreted directly, allowing for greater understanding of the interaction between the features and domain, identity, and expression factors. The BetaMix method has shown promising results across multiple applications, including feature selection and classification [52].

## 1.2 Contributions

We propose novel deep domain adaptative FACial Expressions fusing BEtaMix SElected Landmark Features (FACE-BE-SELF) for domain-invariant expression classification. To the best of our knowledge, our proposed deep domain adaptive FACE-BE-SELF approach is the first to perform concurrent adult-child domain adaptation and learn a generalized expression representation that may be used for both child and adult facial expression classification. Our contributions are as follows:

- We fuse facial landmark measurements with deep feature representations for robust expression learning across age groups.
- Our facial landmark features are decomposed based on expression, domain, and identity correlations.
- A novel statistical method based on a mixture of Beta distributions is proposed for facial feature selection for deep learning.
- A new variant of concurrent adult to child expression learning is performed to yield domain-invariant facial expression classification.
- The proposed domain adaptation method is compared to baseline CNN, transfer learning, and existing domain adaptation methods for facial expression recognition using multiple benchmark data sets.

The remainder of this paper is organized as follows. Section 2 describes the methodology of our approach. Section 3 and 4 present the results and discussion, respectively. Section 5 discusses limitations and Section 6 presents the conclusion.

## 2 METHODS

### 2.1 Data sets

We evaluate our proposed method using four data sets of facial expression images: 1) the Extended Cohn-Kanade (CK+) data set [21], [22], 2) the Aff-Wild2 data set [53], [54], [55], [56], [57], [58], [59], 3) the Child Affective Facial Expression (CAFE) data set [29], [30], and 4) the Child Emotion Facial Expression Set (ChildEFES) [31]. We consider both posed and spontaneous data sets. While spontaneous data sets represent most expressions seen in daily life, posed expressions serve a valuable purpose in healthcare applications such as social reciprocity training [60], [61], [62] for individuals with autism and facial rehabilitation exercises for individuals with Parkinson's disease and facial palsy [63], [64].

### 2.1.1 CK+ data set

The CK+ data set [21], [22] consists of 593 image sequences of posed facial expressions, including labeled 'anger', 'disgust', 'fear', 'happy', 'sad', 'surprise', and 'contempt' examples, captured from 123 adult subjects (ages 18 to 50 years). A mixture of color and grayscale sequences are present in the data set. Sequences vary in length, but each sequence begins with the neutral expression and ends with the peak expression frame, which has been coded for action units (AUs) from the facial action coding system (FACS). We assign the last three frames of a sequence with its corresponding expression label and label the first frame of each sequence as 'neutral'. This yields 1254 samples: 135 'anger', 177 'disgust', 75 'fear', 207 'happy', 327 'neutral', 84 'sad', and 1369 'surprise'.

### 2.1.2 Aff-Wild2 data set

The Aff-Wild2 data set [53], [54], [55], [56], [57], [58], [59], an extension of the Affect-in-the-Wild (Aff-Wild) [65], [66], [67] data set, consists of 558 YouTube videos with annotations for three behavioral tasks: valance and arousal, FACS AUs, and facial expressions ('anger', 'disgust', 'fear', 'happy', 'neutral', 'sad', 'surprise', and 'other'). The facial expression subset of Aff-Wild2 contains 84 videos with 84 ethnically diverse subjects (42 female). Age labels are not provided. Visually, the subjects appear to be mostly adults with few child subjects, including infants. Labeled frames show a variety of different head poses, occlusions, and illumination conditions. Excluding the frames labeled 'other', there are a total of 451794 samples: 18940 'anger', 14545 'disgust', 11336 'fear', 97862 'happy', 197314 'neutral', 80517 'sad', and 31280 'surprise'.

### 2.1.3 CAFE data set

The CAFE data set [29], [30] consists of 1192 color photographs of 154 child subjects (ages 2 to 8 years) posing 'anger', 'disgust', 'fear', 'happy', 'sad', and 'surprise' expressions, including 'neutral'. The data set includes open and closed mouth variations for each expression except 'surprise', which is posed with open mouth only. We include the mouth closed variant of all expressions except for 'surprise', yielding 707 samples: 119 'anger', 96 'disgust', 79 'fear', 120 'happy', 129 'neutral', 62 'sad', and 102 'surprise'. The data usage agreement for the CAFE dataset does not allow republication of the images.

### 2.1.4 ChildEFES data set

The ChildEFES data set [31] consists of color photos and videos capturing 34 child subjects (ages 4 to 6 years) producing a mixture of spontaneous and posed 'anger', 'disgust', 'fear', 'happy', 'sad', 'surprise', and 'contempt' expressions. The expression labels are assigned based upon the agreement of four FACS judges. We crop the expression-labeled videos to the peak expression. Then, we sample the cropped videos at 20 frames per second to generate image sequences. Since the photographs are a subset of the image sequences, we include only the frames sampled from the videos. This yields 9420 (5107 spontaneous) samples: 1435 (170) 'anger', 1196 (468) 'disgust', 655 (19) 'fear', 2196 (1535) 'happy', 2445 (2372) 'neutral', 1053 (450) 'sad', and 440 (93) 'surprise'. The data usage agreement for the ChildEFES data set does not allow for the use of the images in publications.

### 2.1.5 Notation

Let input space $\mathcal{X}$ represent the set of all possible facial images and features. Output space $\mathcal{Y} = \{0, ..., K-1\}$ is the set of $K$ expression class labels ('anger', 'disgust', 'fear', 'happy', 'neutral', 'sad', 'surprise'). $\mathcal{X}$ and $\mathcal{Y}$ are related by a function $f: \mathcal{X} \to \mathcal{Y}$. We consider adult facial expressions (CK+, Aff-Wild2) as the source domain and child facial expressions (CAFE, ChildEFES) as the target domain. We represent each source data set as $D_S = \{(x_i^S, y_i^S) \mid x_i^S \in \mathcal{X}, y_i^S \in \mathcal{Y}\}_{i=1}^{N_S}$, $x_i^S \sim p_X^S$ where $N_S$ is the total number of samples and $p^S$ is the source probability distribution. We represent each target dataset as $D_T = \{(x_i^T, y_i^T) \mid x_i^T \in \mathcal{X}, y_i^T \in \mathcal{Y}\}_{i=1}^{N_T}$, $x_i^T \sim p_X^T$ where $N_T$ is the total number of samples and $p^T$ is the target probability distribution.

## 2.2 Preprocessing

Data sets are preprocessed following [26]. The dlib (http://dlib.net/) library is used to detect the face in each image and extract landmark coordinates on the face. The landmarks are used to center and rotate the face so that the eyes are level. The images are cropped such that the left eye is located 30% of the image width in pixels from the left edge. Images are resized to 256 by 256 pixels, converted to grayscale, and normalized to range [0, 1].

## 2.3 Feature extraction

Using the dlib library, we extract landmark points located at and around facial features such as the nose, eyes, mouth, and eyebrows as well as the perimeter of the face. These landmark locations are used to derive geometric features from FEA and AIFR literature based on pairs and triplets of landmarks. Inter-landmark distance features [36], [44], [45], [49] are measured as the Euclidean distance between pairs of landmarks. Facial triangles [43], [44], [48] are extracted based on a Delaunay triangulation over the landmark locations. Each triangle is represented by a landmark triplet and has four associated features: the area of the triangle and its three angles expressed in radians. Fig. 1 shows examples of the extracted features.

## 2.4 Landmark feature decomposition and selection

We fit the BetaMix method [52] to find significant correlations between the extracted features from adult-child data and three experimental factors taken from the labeled data: expression, domain, and identity. Based on given data, the BetaMix method automatically learns a threshold that represents significant correlations among pairwise features and factors. The extracted features for the source and target data sets are concatenated to form a matrix of $P$ predictors and $N$ samples, where $P > N$. Expression, domain, and identity are also entered into the model, yielding a $(P + 3) \times N$ matrix. We assume the data as $P + 3$ points in $\mathbb{R}^N$. Subspaces of $\mathbb{R}^N$ lie on the Grassmann manifold, a special type of Riemannian manifold with a nonlinear structure [52]. The Grassmann manifold $\mathbb{G}_{N,d}$ is used to study $d$-dimensional subspaces of $\mathbb{R}^N$ [52]. Principal angles $(\theta_1, \ldots, \theta_d)$ between $d$ and $l$-dimensional subspaces in $\mathbb{G}_{N,d}$ have an invariant measure for $d \leq l$ that can be used to compute the volume and probability of their sets [52]. These principal angles can be used to determine canonical correlations $(\rho_1, \ldots, \rho_d)$ as $\rho_k = \cos \theta_k$ with pairs of canonical variables $\{\varphi_k, \phi_k\}_{k=1}^d$ where $\varphi_k \in \mathbb{G}_{N,d}$, $\phi_k \in \mathbb{G}_{N,l}$ [52]. When $d = l = 1$, $\mathbb{G}_{N,1}$ corresponds to lines through the origin of Euclidean space [52]. The line is a natural choice of projection due to its computational ease and interpretability. Furthermore, when $d = l = 1$, the random variable $\sin^2 \theta$ has the following Beta distribution [52]:

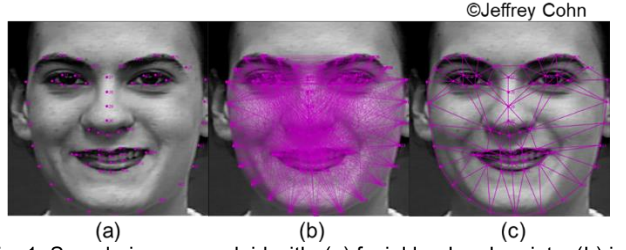

Fig. 1. Sample image overlaid with: (a) facial landmark points, (b) inter-landmark distance features, (c) Delaunay triangulation of the face.

$$\lambda \stackrel{\text{def}}{=} \sin^2 \theta \sim Beta\left(\frac{N-1}{2}, \frac{1}{2}\right), \quad (1)$$

Thus, we consider that the predictors and factors lie on $\mathbb{G}_{N,1}$ and define $\theta_k$ as the angle between the $k$th pair of predictors/factors, $k = 1, \ldots, ((P+3)(P+2))/2$ [52]. We let $\lambda_k = \sin^2 \theta_k$. A predictor or factor is considered 'null' if it corresponds to a randomly sampled point in $\mathbb{R}^N$. As shown in [52], pairs of null predictors/factors are expected to be mutually perpendicular with high probability, even for moderate values of $N$. In relation to Equation (1), a mixture of Beta distributions may be used to determine if the pair represented by each $\lambda_k$ are 'null' (uncorrelated) or 'nonnull' (correlated). Then, the BetaMix model is defined as:

$$l(\lambda_k) = \iota_{0_k} d_0(\lambda_k) + (1 - \iota_{0_k})d(\lambda_k), \quad (2)$$

where $d_0(\lambda_k)$ is the null distribution, $d(\lambda_k)$ is the alternative distribution, $\iota_{0_k} \sim Ber(p_0)$ is a random indicator that equals one if the $k$th pair of predictors corresponding to $\lambda_k$ are 'null', and $p_0$ is the probability of the 'null' component. The 'null' component of the mixture model is defined by the Beta distribution:

$$d_0(\lambda_k) = \frac{1}{Beta\left(\frac{s-1}{2}, \frac{1}{2}\right)} \lambda_k^{\frac{s-1}{2}-1}(1-\lambda_k)^{-\frac{1}{2}}, \quad (3)$$

where $s \leq N$ is the estimated effective sample size. The 'nonnull' component of the mixture model is defined as:

$$d(\lambda_k) = \frac{1}{Beta(\alpha, \beta)} \lambda_k^{\alpha-1}(1-\lambda_k)^{\beta-1}, \quad (4)$$

where $\alpha, \beta > 0$. The latent mixture variables $(\alpha, \beta, s)$ are estimated using the EM algorithm.

The E-step updates $\iota_{0_k}$ with the posterior mean:

$$\hat{\iota}_{0_k} = \frac{p_0 d_0(\lambda_k)}{p_0 d_0(\lambda_k) + (1-p_0)d_0(\lambda_k)} \quad (5)$$

and $p_0$ is updated with its maximum likelihood estimate, $\hat{p}_0 = \mathbb{E}(\hat{\iota})$. The M-step obtains the maximum likelihood estimates of $\alpha, \beta,$ and $s$ by solving the following equations:

$$\psi(\alpha) - \psi(\alpha + \beta) = \frac{\sum_{k=1}^{(P+3)(P+2)/2}(1-\iota_{0_k})\log(\lambda_k)}{\sum_{k=1}^{((P+3)(P+2))/2}(1-\iota_{0_k})} \quad (6)$$

$$\psi(\beta) - \psi(\alpha + \beta) = \quad (7)$$
$$\frac{\sum_{k=1}^{(P+3)(P+2)/2}(1-\iota_{0_k})\log(1-\lambda_k)}{\sum_{k=1}^{(P+3)(P+2)/2}(1-\iota_{0_k})}$$

$$\psi\left(\frac{s-1}{2}\right) - \psi\left(\frac{s}{2}\right) = \frac{\sum_{k=1}^{(P+3)(P+2)/2}\iota_{0_k}\log(\lambda_k)}{\sum_{k=1}^{(P+3)(P+2)/2}\iota_{0_k}} \quad (8)$$

where $\psi(\cdot)$ is the digamma function. The E- and M-steps are repeated iteratively to update the parameters until convergence. Pairs of predictors/factors are considered

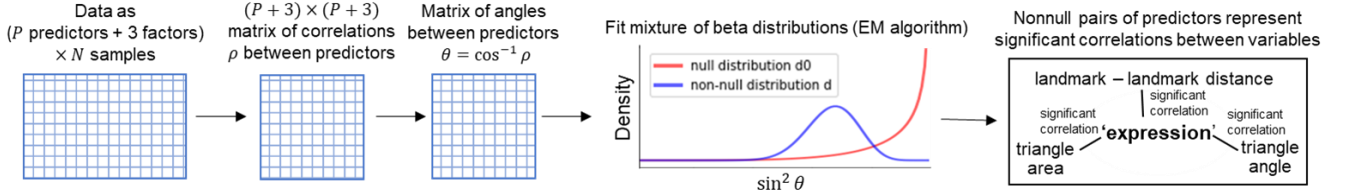

Fig. 2. Overview of the BetaMix method. Data is entered as a matrix of $P$ predictors + 3 factors and $N$ samples. Correlations $\rho$ among predictors/factors are computed and $\theta = \cos^{-1} \rho$ is used to compute the pairwise angles between predictors from the correlations. Next, a mixture of Beta distributions is fit over random variables $\sin^2 \theta$. Pairs of predictors are considered 'nonnull' if the posterior null probability under $d_0$ is smaller than a threshold $\tau$. These 'nonnull' pairs represent significant correlations between variables. The 'nonnull' pairs for expression, domain, and identity are used to decompose the feature vector into sets correlated with each of these three factors.

'nonnull' if the posterior null probability under $d_0$ is smaller than threshold $\tau$, $\hat{\iota}_{0_k} < \tau$. We denote the maximum $\lambda_k$ that satisfies $\hat{\iota}_{0_k} < \tau$ as $Q$. Then, the screening rule for nonnull pairs may be written as $\lambda_k < Q$. Since $\lambda_k = \sin^2 \theta_k$ and $\rho_k = \cos \theta_k$, pairs with a correlation of at least $\rho = \cos(\sin^{-1}(Q^{1/2}))$ are considered significant. Fig. 2 summarizes the BetaMix method.

Based on the fitted Beta distribution, a graphical model is built where each node is a predictor or factor. An edge connects each nonnull predictor-predictor pair or factor-predictor pair. These edges represent a significant correlation between the connected nodes (predictors or factors). A subgraph formed by a factor and its adjacent predictor nodes captures the subset of predictors that are significantly correlated with the factor. Using these subgraphs, we decompose the feature vector into sets correlated with expression, domain, and identity. For our proposed FACE-BE-SELF approach, we select the features in the expression subgraph and prune features that also appear in the domain or identity subgraphs. The resulting selection of features is used in subsequent feature fusion.

### 2.5 Deep Learning Models

We model supervised classification as the following inverse problem:

$$Y = f(X; w) \quad (9)$$

where $f(\cdot)$ is a neural network model parameterized by weights $w$, $X \in \mathcal{X}$ are the model inputs, and $Y \in \mathcal{Y}$ are the associated class labels. We partition $f(\cdot)$ into feature extractor $M: \mathcal{X} \to \mathcal{Z}$ and classifier $C: \mathcal{Z} \to \mathcal{Y}$ such that $f = C \circ M$ with latent feature space $\mathcal{Z}$. Using this notation, we define multiple architectures: a multi-layer perceptron (MLP), CNN, and feature fusion model including MLP and CNN components.

For the MLP, we consider $X = V$, where feature set $V \in \mathcal{V} = \mathbb{R}^{P_{Beta}}$ and $P_{Beta}$ is the number of BetaMix-selected features based on significant correlations with expression. The MLP has one hidden layer with 512 hidden units, ReLU activation, and dropout with a probability of 0.5. We consider a latent feature vector $Z \in \mathcal{Z} = \mathbb{R}^{512}$ produced by the hidden layer of the MLP. The hidden layer is followed by a softmax output layer of $K = (number\ of\ classes)$ nodes. Uniform initialization is applied to all of the MLP weights.

For the CNN, we consider $X = U \in \mathcal{U} = \mathbb{R}^{256 \times 256}$ and define $M(\cdot)$ as a sequence of three convolutional blocks, each consisting of a convolutional layer with 3x3 filter kernels followed by a 2x2 maximum response pooling, and a fully connected neural network with 512-dimensional hidden layer. This hidden layer also yields a latent feature vector $Z \in \mathcal{Z} = \mathbb{R}^{512}$. We use the uniform distribution to initialize all weights. Dropout with a probability of 0.5 is applied to the 512-dimensional hidden layer. We define $C(\cdot)$ as a $K$-dimensional fully connected layer with softmax mapping from $\mathcal{Z}$ onto $\mathcal{Y}$. This CNN architecture is shown in Fig. 3.

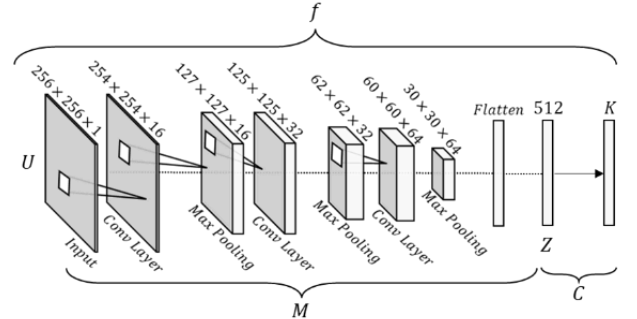

Fig. 3. CNN architecture. Model $f(\cdot)$ is partitioned into feature extractor $M(\cdot)$, mapping from input $U \in \mathbb{R}^{256 \times 256}$ to latent feature vector $Z$, and classifier $C(\cdot)$, mapping from $Z$ to the $K$-dimensional output.

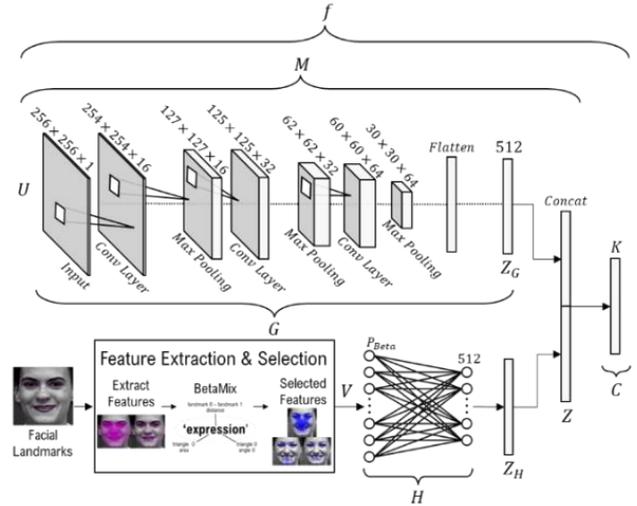

Fig. 4. Feature fusion architecture. Model $f(\cdot)$ is partitioned into feature extractor $M(\cdot)$ and classifier $C(\cdot)$. $M(\cdot)$ consists of CNN model $G: \mathcal{U} \to \mathcal{Z}_G$ and MLP $H: \mathcal{V} \to \mathcal{Z}_H$. $Z_G$ and $Z_H$ are concatenated as $Z$ and classifier $C(\cdot)$, maps from $Z$ to the $K$-dimensional output.

For our proposed FACE-BE-SELF feature fusion model, we define $X$ as a tuple $(U, V)$, where $U \in \mathcal{U} = \mathbb{R}^{256 \times 256}$ and $V \in \mathcal{V} = \mathbb{R}^{P_{Beta}}$. Feature extractor $M(\cdot)$ is made up of CNN model $G: \mathcal{U} \to \mathcal{Z}_G$, $\mathcal{Z}_G = \mathbb{R}^{512}$ and the MLP model $H: \mathcal{V} \to \mathcal{Z}_H$, $\mathcal{Z}_H = \mathbb{R}^{512}$. We define the concatenation of $\mathcal{Z}_G$ and $\mathcal{Z}_H$ spaces as $Z \in \mathcal{Z} = \mathbb{R}^{1024}$. Then, we define $C(\cdot)$ as a $K$-dimensional fully connected layer with softmax mapping from $\mathcal{Z}$ onto $\mathcal{Y}$. The architecture of the feature fusion model

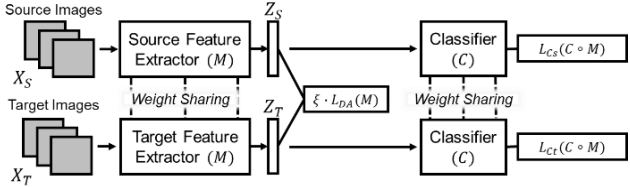

Fig. 5. Domain adaptation framework. Source-target pairs $(X_S, X_T)$ are passed into parallel feature extractors $M(\cdot)$. Resulting latent distributions $Z_S$ and $Z_T$ are aligned by domain alignment loss $\mathcal{L}_{DA}$. Parallel classifiers $C(\cdot)$, are supervised by source and target classification losses $\mathcal{L}_{Cs}$ and $\mathcal{L}_{Ct}$, respectively.

is shown in Fig. 4.

## 2.6 Deep Domain Adaptation

Rather than maximizing the performance on a target domain, our goal for deep domain adaptation is to optimize the model for maximum performance on both the source and target domains. We assume that the distribution shift between source and target domains can be attributed to covariate shift $p_X^S(x) \neq p_X^T(x)$, rather than a shift in the label distributions, i.e. we assume $\forall x \in \mathcal{X}, p^S(Y|X = x) = p^T(Y|X = x)$. We adopt a dual stream architecture (Fig. 5) consisting of parallel feature extractors $M_S(\cdot)$ and $M_T(\cdot)$ for source and target distributions, respectively. Weights are shared between the two branches such that $M(\cdot) = M_S(\cdot) = M_T(\cdot)$. Paired source and target examples $X_S$ and $X_T$ are passed into their respective feature extractors to yield source and target latent representations, i.e., $Z_S = M(X_S)$ and $Z_T = M(X_T)$. Parallel classifiers $C(\cdot)$, which also share weights, are trained with $Z_S$ and $Z_T$ to optimize performance on both source and target domains.

The model is optimized using three supervised loss functions: source classification loss $\mathcal{L}_{Cs}(f)$, target classification loss $\mathcal{L}_{Ct}(f)$, and domain alignment loss $\mathcal{L}_{DA}(M)$. We define $\mathcal{L}_{Cs}$ and $\mathcal{L}_{Ct}$ as the categorical cross-entropy loss given our multiclass expression classification problem. To address class imbalance in the training sets, we scale each sample's contribution to the overall loss by the frequency of its associated class in the training set. We define $\mathcal{L}_{DA}$ as the contrastive alignment loss [38]:

$$\mathcal{L}_{DA}(M) = \sum_{a=1}^{K}\sum_{i,j} d\left(M(x_i^S|y_i^S = a), M(x_j^T|y_j^T = a)\right) \\ + \sum_{a,b|a \neq b}^{K}\sum_{i,j} k\left(M(x_i^S|y_i^S = a), M(x_j^T|y_j^T = b)\right) \quad (10)$$

We follow [38] in selecting $d(\cdot)$ and $k(\cdot)$ as:

$$d(M(x_i^S), M(x_j^T)) = 0.5 \|M(x_i^S) - M(x_j^T)\|_F^2 \quad (11)$$

and

$$k(M(x_i^S), M(x_j^T)) \\ = \tfrac{1}{2}(max(0, m - \|M(x_i^S) - M(x_j^T)\|_F))^2, \quad (12)$$

where $\|\cdot\|_F$ is the Frobenius norm and margin $m = 1$ [38]. The effect of $\mathcal{L}_{DA}$ is to minimize (1) distance between samples of the same class from different domains, and (2) similarity between samples of different classes and domains. The overall loss is:

$$\mathcal{L} = (\mathcal{L}_{Cs} + \mathcal{L}_{Ct}) + \xi\mathcal{L}_{DA}, \quad (13)$$

where $0 < \xi < 1$ is a scaling parameter for balancing the contribution of domain alignment loss.

## 2.7 Experiments

We perform preprocessing of the CAFE, ChildEFES, CK+, and Aff-Wild2 data sets following Section 2.2. To evaluate our proposed FACE-BE-SELF method, we consider data sets in two source/target pairs: CK+/CAFE (posed expressions only) and Aff-Wild2/ChildEFES (majority spontaneous expressions). We split each data set into multiple train, validation, and test sets using a 5x2 nested cross validation design. In the outer 5-fold cross validation loop, the data is split into train (4 folds) and test (1 fold) sets. In the inner loop, the train set is divided into 2 to yield inner train and validation sets for hyperparameter selection. The validation performance metrics are averaged across the two folds to yield the best hyperparameters. These hyperparameter selections are then used to train the model with the recombined outer loop train set and evaluate on the held-out test fold. This procedure is repeated a total of 5 times, such that each sample appears in one of the 5 test sets. To avoid inflation of performance estimates based on subject-specific features, we generate the cross validation folds such that each subject appears in one fold only and no subject appears in both train and test sets [26], [27].

We fit the BetaMix method on the train sets for each source/target pair. The fitted mixture model identifies nonnull (significantly correlated) pairs of predictors/factors which are used to build a graph with predictors/factors represented as nodes and significant correlations represented as edges. By examining the subgraphs of each factor node and its adjacent predictor nodes, we report the mean number of significantly correlated features for each factor and the overlap of features appearing in multiple factor subgraphs. To select features for subsequent fusion, we consider the expression subgraph, pruning features that also appear in the domain and/or subject subgraphs. We assess the discriminability of our data-driven feature selection compared to that of features selected based upon a range of correlation thresholds (0.1, 0.2, …, 1.0).

The average overall F1 performance on the inner 2-fold cross validation loop is used to select a value for the loss balancing parameter $\xi$ (Equation (13)) for each of the outer 5-fold cross validation train sets. Other studies [68], [69] have found that $\xi$ is problem-specific and consider values in the range (0.00, 1.00). Due to high computational costs, we choose among representative low (0.01), moderate (0.3), and high (0.8) values in (0.00, 1.00). To better understand the contributions of CNN, BetaMix-selected landmark features, and domain adaptation to our proposed FACE-BE-SELF approach, we perform an ablation study.

Then, we evaluate the performance of our proposed domain adaptation with FACE-BE-SELF approach on two source/target data set pairs and compare against four baseline models: 1) CNN trained on source data (source CNN) [26], 2) CNN trained on target data (target CNN) [26], 3) three transfer learning approaches (pretraining on source data then, a, training on target data [26], b, fine-tuning on the target data [26], or c, fine-tuning on a mixture of source and target data), and 4) two existing domain

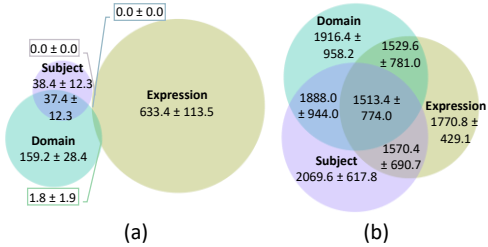
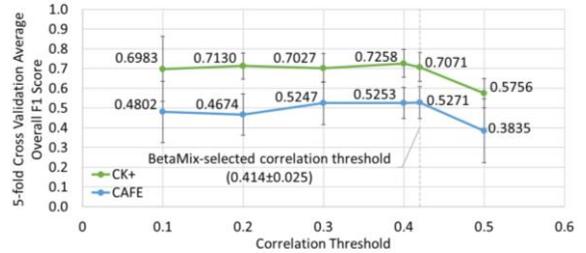

Fig. 6. Mean number of features correlated with expression, domain, and identity for (a) CK+/CAFE and (b) Aff-Wild2/ChildEFES.

Fig. 7. CK+/CAFE 5-fold cross validation average overall F1 scores for MLP trained on expression-correlated feature selections at various thresholds.

TABLE 1
ABLATION STUDY FOR THE PROPOSED FACE-BE-SELF MODEL

| Model Description | Included Model Components | | | Overall F1 Score | |
|---|---|---|---|---|---|
| | CNN | BetaMix-Selected Landmark Features | Domain Adaptation | Source (CK+) | Target (CAFE) |
| MLP with Betamix-selected landmark features, trained on source | | ✓ | | 0.7071 ± 0.0640 | 0.3221 ± 0.0996 |
| CNN trained on source | ✓ | | | 0.8119 ± 0.0546 | 0.4713 ± 0.0965 |
| CNN fusing Betamix-selected landmark features, trained on source | ✓ | ✓ | | 0.8358 ± 0.0385 | 0.4705 ± 0.0479 |
| MLP with Betamix-selected landmark features, domain adaptation | | ✓ | ✓ | 0.6514 ± 0.0864 | 0.4138 ± 0.0540 |
| CNN, domain adaptation | ✓ | | ✓ | 0.8409 ± 0.0476 | 0.7765 ± 0.0171 |
| **Ours** | ✓ | ✓ | ✓ | **0.8443 ± 0.0466** | **0.8303 ± 0.0286** |

adaptation approaches [27], [38].

For all experiments, we train deep models using the ADAM optimizer with a triangular learning rate policy [70] cycling between a minimum learning rate of $\zeta_{min} = 10^{-5}$ and a maximum learning rate of $\zeta_{max} = 10^{-3}$. We use a batch size $n = 32$.

## 3 RESULTS

### 3.1 Feature extraction

We extract 68 landmark points on the face as shown in Fig. 1(a) and use these to measure inter-landmark distances. Because the 68 × 68 Euclidean distance matrix is symmetric with zeros (self-distance) in diagonal entries, the total number of inter-landmark distance features is (68 $landmarks$ × 68 $landmarks$) − 68/2 = 2278. Fig. 1(b) overlays all possible inter-landmark distance features on the face. The Delaunay triangulation over the landmark locations results in a set of 106 triangles on the face. For each facial triangle, the area and three internal angles are computed, resulting in 106 $triangles$ × (4 $features/triangle$) = 424 triangle-based features. Fig. 1(c) visualizes the Delaunay triangulation on the face.

### 3.2 Selection of landmark features for expression, domain, and identity factors

We fit the BetaMix method on each of the 5-fold cross validation training sets for both source/target data set pairs. For a representative CK+/CAFE training set, BetaMix learns the screening rule $\lambda_k = \sin^2(\theta) < Q = 0.83$. This is equivalent to an angle of 65.7° or less between the pairs of factors/features, or a correlation coefficient of at least $\rho = \cos(65.7°) = 0.412$. Averaging over the 5 training sets, the mean correlation threshold for CK+/CAFE is 0.414±0.025. Similarly, for a representative Aff-Wild2/ChildEFES training set, BetaMix learns the screening rule $\lambda_k = \sin^2(\theta) < Q = 0.99$. This is equivalent to an angle of 84.3° or less between the pairs of factors/features, or a correlation coefficient of at least $\rho = \cos(84.3°) = 0.100$. For Aff-Wild2, the mean correlation threshold is 0.045±0.031. Fig. 6 shows the mean number of features correlated with 'expression', 'domain', and 'identity', as well as the number correlated with two out of three and all three factors. Considering the CK+/CAFE pair of data sets, Fig. 7 compares the performance of an MLP trained on features selected by the data-driven correlation threshold learned by BetaMix and those selected based upon a range of correlation thresholds (0.1, 0.2, …, 1.0). There is not any feature with a correlation coefficient of 0.6 or greater for the expression factor.

### 3.3 Domain Adaptation

For each train/test split in the outer 5-fold cross validation loop, we select $\xi$ based on the overall F1 score averaged over the 2 validation sets of the inner 2-fold cross validation loop. For CK+/CAFE, $\xi = 0.01$ is selected for all 5 train/test splits of the outer cross validation loop. For Aff-Wild2/ChildEFES, $\xi = 0.01$ is selected once, $\xi = 0.3$ is selected twice, and $\xi = 0.8$ is selected twice.

Ablation study results for FACE-BE-SELF are presented in Table 1. The proposed model is compared with variants that selectively remove one or two of the following model components: CNN, BetaMix-selected landmark features, and domain adaptation. Fig. 8 compares the 5-fold cross validation performance of our proposed FACE-BE-SELF with multiple baselines for the CK+/CAFE and Aff-Wild2/ChildEFES source/target pairs, including CNNs trained on a single domain [26], transfer learning [26], and domain adaptation approaches [27]. Since transfer learning performance on the source domain is expected to deteriorate after fine-tuning on target data only, we also compare with transfer learning fine-tuned on a mixture of source and target data. Fig. 9 plots one-versus-rest multiclass receiver operating characteristic (ROC) curves and reports the associated area under curve (AUC) metrics for the proposed FACE-BE-SELF approach.

## 4 DISCUSSION

This paper presents FACE-BE-SELF for classification of adult and child facial expressions through deep domain adaptation and the fusion of facial landmark features correlated with expressions. Our experiments on four data sets and comparison of eight facial expression classification methods have revealed four important findings as follows. First, the decomposition of landmark features for expression, domain, and identity factors based on the data-driven threshold learned by BetaMix reveals very little overlap in the subgraphs of different factors for CK+/CAFE (Fig. 6 (a)) while the factor subgraphs of Aff-Wild2/ChildEFES share a substantial number of adjacent feature nodes (Fig. 6 (b)). Features concurrently correlated with expression and domain factors indicate the presence of domain shift in the landmark feature space $\mathcal{V} = \mathbb{R}^{P_{Beta}}$. While the CNN feature space is known to exhibit adult-child domain shift [25], [26], [27], [28], our results suggest the domain shift to be dependent on the domain data set pair. The underlying data dependency (differences in overlap regions of Fig. 6 (a) and Fig. 6 (b)) may be attributed to differences in sample size and demographics, age ranges, and/or mixture of posed/spontaneous expressions [16], [17]. Second, a parsimonious feature selection is obtained from the expression subgraph after eliminating features significantly correlated with the other factors (Fig. 7). Third, our ablation study shows that fusing these selected landmark features and CNN-extracted image features improves the expression classification performance for both child and adult data (Table 1). Fourth, our proposed FACE-BE-SELF method outperforms all baseline models for the posed data sets (CK+/CAFE) and performs competitively for the data sets with spontaneous expressions (Aff-Wild2/ChildEFES). The sections to follow provide detailed discussions in addition to and expanding upon these four key findings.

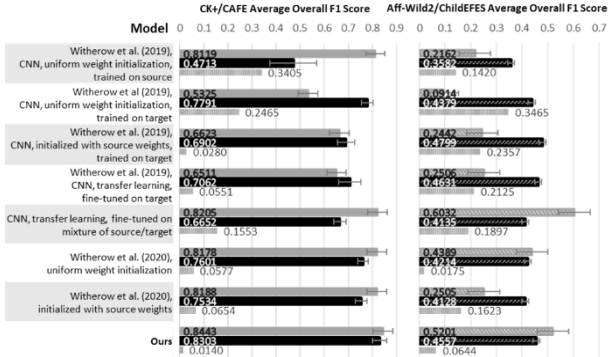

Fig. 8. 5-fold cross validation overall F1 score for comparison models.

### 4.1 Selection and fusion of facial landmark features

Our comparison of the proposed data-driven feature selection and a range of correlation thresholds reveals that our data driven BetaMix approach yields the largest correlation threshold prior to substantial performance degradation (Fig. 7). This threshold corresponds to a parsimonious selection of highly correlated features that preserve useful complementary information for expressions that is discarded at higher thresholds. Furthermore, fusing CNN-extracted features with the selected landmark features improves the classification performance of child and adult facial expressions (Table 1). Like age estimation and AIFR, facial expression classification also benefits from the fusion of geometric landmark and texture features [39], [40]. Given that the feature fusion model outperforms CNN features only, our selected landmark features provide complementary information representative of expressions beyond that learned by the CNN (Table 1). The effectiveness of selected features in classification suggests that the proposed BetaMix correlation coefficient threshold is an effective metric in optimizing feature selection for facial expression classification.

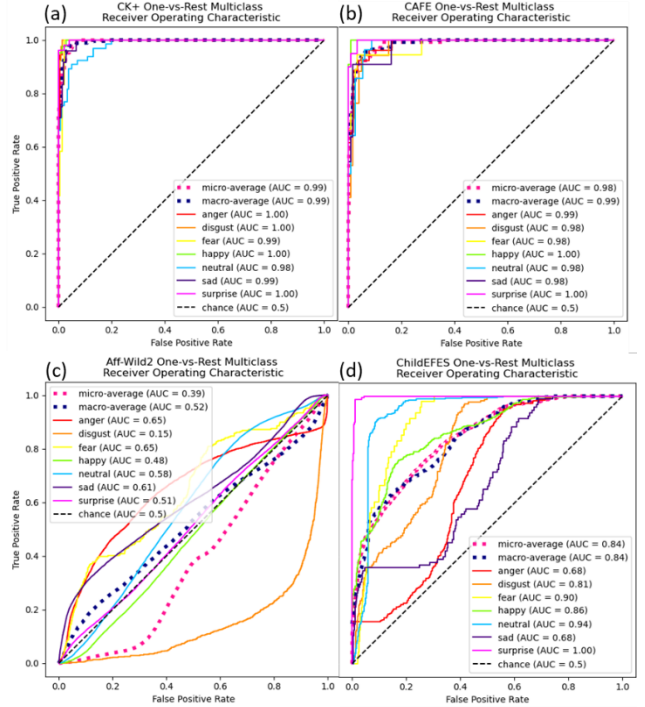

Fig. 9. FACE-BE-SELF ROC Curves for various data sets.

### 4.2 Domain adaptation for expression learning

Our findings suggest that domain adaptation methods provide robust representation learning of adult and child facial expressions (Fig. 8). During adaptation, source and target performance are jointly optimized via $\mathcal{L}_{Cs}$ and $\mathcal{L}_{Ct}$ while the class conditional distributions are aligned using $\mathcal{L}_{DA}$. This optimization procedure ensures balanced performance on both domains. Our findings also confirm that supervision on both domains (as in transfer learning) or a method of domain alignment is required for effective classification (Fig. 8). For both CK+/CAFE and Aff-Wild2/ChildEFES source-target pairs, we observe poor cross domain performance for CNNs trained on a single domain (Fig. 8). This poor cross domain performance is indicative of distribution shift and replicates the findings of multiple prior studies [26], [27], [28].

Our proposed FACE-BE-SELF method yields higher source and target average overall F1 scores for CK+/CAFE than all baseline models, with similar average overall F1 scores for source and target of 0.8443 and 0.8303, respectively, with a difference of 0.0140 (Fig. 8). Spontaneous expression classification (Aff-Wild2/ChildEFES) is more challenging than classification of posed facial expressions such as CK+ and CAFE. For example, Aff-Wild2 is the most challenging of the four data sets that we use to evaluate our approach. Current state-of-the-art performance on the official test set for Aff-Wild2 is an overall F1 score of 0.3587, achieved by the best performing team at the recent 3rd Affective Behavior Analysis in-the-wild Competition [71]. Please note that our results are not directly comparable as we perform cross validation rather than use the official test set. For Aff-Wild2/ChildEFES, the best performing models are FACE-BE-SELF and fine-tuning on a mixture of source and target data (Fig. 8). Compared to fine-tuning on a mixture of source and target data, FACE-BE-SELF performs better on ChildEFES (average overall F1 score 0.4557 > 0.4214) and worse on Aff-Wild2 (average overall F1 score 0.5201 < 0.6032) but has a smaller difference in source and target performance (0.0644 vs 0.1897). Thus, despite poorer performance on Aff-Wild2, FACE-BE-SELF offers better target (ChildEFES) performance and more balanced performance between source and target.

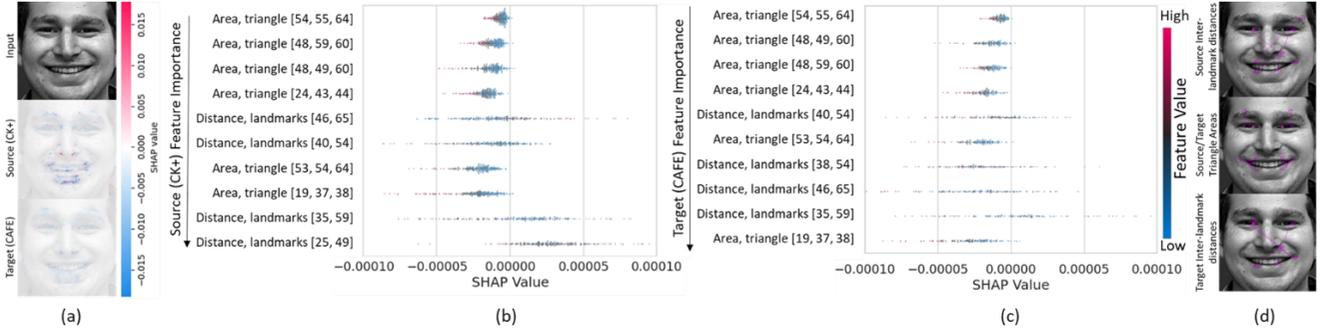

Fig. 10. Visualization of SHAP values for FACE-BE-SELF: (a) image input to CNN, (b) top 10 source landmark features, (c) top 10 target landmark features, (d) plotted top 10 landmark features for source and target.

The ROC curves for CK+/CAFE (Fig. 9 (a)(b)) reveal that despite class imbalance during training, FACE-BE-SELF learns to recognize all classes with AUCs near unity, indicating high sensitivity and specificity. For ChildEFES, the ROC curves show that all classes perform better than chance (Fig. 9 (d)). 'Surprise', with its distinctive open mouth appearance achieves an AUC of unity while negative expressions 'anger' and 'sad' prove more difficult. The ROC curves for Aff-Wild2 (Fig. 9 (c)) reflect the challenging nature of the data set with an overall average AUC of 0.52, close to chance level (0.50). The best performing classes are 'anger' (AUC 0.65) and 'fear' (AUC 0.65), while 'disgust' performs worst (0.15).

### 4.4 Explaining feature contributions

We perform additional analysis using SHapley Additive exPlanations (SHAP) [72] to explain the contributions of different features to the classification of child and adult expressions. To quantify the contributions of both BetaMix-selected landmark features and CNN features in the feature fusion model, we use the expected gradients method [73] as implemented in the SHAP library (https://github.com/slundberg/shap) to obtain and visualize the (approximate) SHAP values for both landmark and CNN features. Fig. 10 visualizes the SHAP values for source (CK+) and target (CAFE) domains. We use the same image from the CK+ data set for all visualizations. Fig. 10 (a) shows the SHAP values associated with source and target image inputs to the CNN feature extractor. For both source and target, areas of the input with the greatest (positive or negative) contribution to expression classification are those involved in producing facial expressions: the eyebrows, eyes, nose, and mouth. Fig. 10 (b) and Fig 10 (c) visualize the SHAP values of the top ten most important landmark features for source and target, respectively, ranked based on their mean absolute SHAP value. Fig. 10 (d) plots these top ten features. Nine out of the ten features are the same for the source and target sets. The top four features, which are ranked in the same order for both source and target, are areas of triangles located at the right corner of the lips (2 features), left corner of the lips (1 feature), and between the left eye and eyebrow (1 feature). The symmetric features (the second area at the left corner of the lips and area between the right eye and eyebrow) are also among the top ten most important features, but in different orders of importance. In addition to these six triangle area features, three inter-landmark distance features are ranked among the top ten for both domains. These features represent distances between the mouth and eyes (2 features) and the mouth and nose (1 feature). As with the image input, the top ten landmark features represent important areas of the face for producing expressions: the eyebrows, eyes, and mouth.

## 5 LIMITATIONS

Although the BetaMix method is robust to dependence among samples, the high degree of similarity among faces (compared to other types of data) and universality of expressions may yield a small effective sample size. Even with a small effective sample size, BetaMix is shown to capture significantly correlated landmark features. However, there may be features that are useful for classification of expressions but are not significantly correlated with expression based on the BetaMix-learned minimum correlation coefficient. Furthermore, the data dependency of BetaMix feature selections may affect performance on unseen data sets. An additional adaptation or fine-tuning step may be required for these models to address possible data dependency. The age ranges studied cover 2 to 8 years for CAFE, 4 to 6 years for ChildEFES, and 18+ years for CK+. Aff-Wild2 does not report specific age ranges. Further research is required to determine if the adapted models are capable of generalizing to participants in other age groups, e.g., teens and pre-teens.

## 6 CONCLUSION

In this work, we propose for the first time in literature novel deep domain adaptative FACE-BE-SELF for concurrent learning of adult and child facial expressions. FACE-BE-SELF yields a meaningful and effective selection of features that are correlated with expressions. The explainability and visualization of SHAP values corroborate the facial expression classification performance of our method. The superiority of our method over existing transfer learning and domain adaption methods satisfies the need for a systematic feature selection, feature fusion, and domain adaptation to perform domain-invariant classification. In future work, we plan to investigate the generalizability of this approach to other age groups and data acquisition pipelines. We hope that this approach may be used to yield automated, objective assessments of age or domain varying patterns in other applications.

## ACKNOWLEDGMENT

This material is based upon work supported by the National Science Foundation Graduate Research Fellowship under Grant No. 1753793 and by the Research Computing clusters at Old Dominion University under National Science Foundation Grant No. 1828593.

## REFERENCES

[1] A. Amodia-Bidakowska, C. Laverty, and P. G. Ramchandani, "Father-child play: A systematic review of its frequency, characteristics and potential impact on children's development," *Developmental Review,* vol. 57, p. 100924, 2020.

[2] H. Chen, H. W. Park, and C. Breazeal, "Teaching and learning with children: Impact of reciprocal peer learning with a social robot on children's learning and emotive engagement," *Computers & Education,* vol. 150, p. 103836, 2020.

[3] M. M. Terwogt and H. Stegge, "Children's perspective on the emotional process," in *The social child*: Psychology Press, 2021, pp. 249-269.

[4] P. Goldberg *et al.*, "Attentive or not? Toward a machine learning approach to assessing students' visible engagement in classroom instruction," *Educational Psychology Review,* vol. 33, pp. 27-49, 2021.

[5] S. K. Gupta, T. Ashwin, and R. M. R. Guddeti, "Students' affective content analysis in smart classroom environment using deep learning techniques," *Multimedia Tools and Applications,* vol. 78, pp. 25321-25348, 2019.

[6] Ö. Sümer, P. Goldberg, S. D'Mello, P. Gerjets, U. Trautwein, and E. Kasneci, "Multimodal engagement analysis from facial videos in the classroom," *IEEE Transactions on Affective Computing,* 2021.

[7] T. Hassan *et al.*, "Automatic detection of pain from facial expressions: a survey," *IEEE transactions on pattern analysis and machine intelligence,* vol. 43, no. 6, pp. 1815-1831, 2019.

[8] G. Zamzmi, R. Paul, D. Goldgof, R. Kasturi, and Y. Sun, "Pain assessment from facial expression: Neonatal convolutional neural network (N-CNN)," presented at the 2019 International Joint Conference on Neural Networks (IJCNN), 2019.

[9] Z. Fei *et al.*, "Deep convolution network based emotion analysis towards mental health care," *Neurocomputing,* vol. 388, pp. 212-227, 2020.

[10] C. Su, Z. Xu, J. Pathak, and F. Wang, "Deep learning in mental health outcome research: a scoping review," *Translational Psychiatry,* vol. 10, no. 1, p. 116, 2020.

[11] K. Owada *et al.*, "Computer-analyzed facial expression as a surrogate marker for autism spectrum social core symptoms," *PLOS ONE,* vol. 13, no. 1, p. e0190442, 2018.

[12] M. D. Samad, N. Diawara, J. L. Bobzien, J. W. Harrington, M. A. Witherow, and K. M. Iftekharuddin, "A Feasibility Study of Autism Behavioral Markers in Spontaneous Facial, Visual, and Hand Movement Response Data," *IEEE Transactions on Neural Systems and Rehabilitation Engineering,* vol. 26, no. 2, pp. 353-361, 2018.

[13] M. D. Samad, N. Diawara, J. L. Bobzien, C. M. Taylor, J. W. Harrington, and K. M. Iftekharuddin, "A pilot study to identify autism related traits in spontaneous facial actions using computer vision," *Research in Autism Spectrum Disorders,* vol. 65, pp. 14-24, 2019.

[14] M. T. Akbar, M. N. Ilmi, I. V. Rumayar, J. Moniaga, T.-K. Chen, and A. Chowanda, "Enhancing game experience with facial expression recognition as dynamic balancing," *Procedia Computer Science,* vol. 157, pp. 388-395, 2019.

[15] P. M. Blom, S. Bakkes, and P. Spronck, "Modeling and adjusting in-game difficulty based on facial expression analysis," *Entertainment Computing,* vol. 31, p. 100307, 2019.

[16] S. Bhattacharya and M. Gupta, "A survey on: Facial emotion recognition invariant to pose, illumination and age," presented at the 2019 Second International Conference on Advanced Computational and Communication Paradigms, 2019.

[17] C. Dalvi, M. Rathod, S. Patil, S. Gite, and K. Kotecha, "A Survey of AI-Based Facial Emotion Recognition: Features, ML & DL Techniques, Age-Wise Datasets and Future Directions," *IEEE Access,* vol. 9, pp. 165806-165840, 2021.

[18] T. Baltrusaitis, A. Zadeh, Y. C. Lim, and L.-P. Morency, "Openface 2.0: Facial behavior analysis toolkit," presented at the 2018 13th IEEE international conference on automatic face & gesture recognition (FG 2018), 2018.

[19] Noldus Information Technology bv. "FaceReader." Noldus Information Technology bv. https://www.noldus.com/facereader (accessed 07/27/2022, 2022).

[20] iMotions A/S. "Facial Expression Analysis." iMotions A/S. https://imotions.com/biosensor/fea-facial-expression-analysis/ (accessed 07/27/2022, 2022).

[21] T. Kanade, J. F. Cohn, and Y. Tian, "Comprehensive database for facial expression analysis," presented at the Proceedings Fourth IEEE International Conference on Automatic Face and Gesture Recognition (Cat. No. PR00580), 2000.

[22] P. Lucey, J. F. Cohn, T. Kanade, J. Saragih, Z. Ambadar, and I. Matthews, "The Extended Cohn-Kanade Dataset (CK+): A complete dataset for action unit and emotion-specified expression," presented at the 2010 IEEE Computer Society Conference on Computer Vision and Pattern Recognition - Workshops, 13-18 June 2010, 2010.

[23] P. Burke and C. Hughes-Lawson, "The growth and development of the soft tissues of the human face," *Journal of anatomy,* vol. 158, p. 115, 1988.

[24] C. Grossard *et al.*, "Children facial expression production: influence of age, gender, emotion subtype, elicitation condition and culture," *Frontiers in psychology,* p. 446, 2018.

[25] A. Dapogny *et al.*, "On Automatically Assessing Children's Facial Expressions Quality: A Study, Database, and Protocol," *Frontiers in Computer Science,* Original Research vol. 1, 2019.

M. A. WITHEROW ET AL.: DEEP ADAPTATION OF ADULT-CHILD FACIAL EXPRESSIONS BY FUSING LANDMARK FEATURES 11
[26] M. Witherow, M. Samad, and K. Iftekharuddin, "Transfer learning approach to multiclass classification of child facial expressions," presented at the SPIE Optical Engineering + Applications, 2019, OPO. [Online].

[27] M. Witherow, W. Shields, M. Samad, and K. Iftekharuddin, "Learning latent expression labels of child facial expression images through data-limited domain adaptation and transfer learning," presented at the SPIE Optical Engineering + Applications, 2020, OPO. [Online].

[28] Z. Zheng, X. Li, J. Barnes, C.-H. Park, and M. Jeon, "Facial Expression Recognition for Children: Can Existing Methods Tuned for Adults Be Adopted for Children?," presented at the International Conference on Human-Computer Interaction, Cham, 2019.

[29] V. LoBue and C. Thrasher. *The Child Affective Facial Expression (CAFE) set*. Databrary, 2014.

[30] V. LoBue and C. Thrasher, "The Child Affective Facial Expression (CAFE) set: validity and reliability from untrained adults," (in English), *Frontiers in Psychology,* Methods vol. 5, 2015.

[31] J. G. Negrão et al., "The Child Emotion Facial Expression Set: A Database for Emotion Recognition in Children," (in English), *Frontiers in Psychology,* Original Research vol. 12, 2021.

[32] R. A. Khan, A. Crenn, A. Meyer, and S. Bouakaz, "A novel database of children's spontaneous facial expressions (LIRIS-CSE)," *Image and Vision Computing,* vol. 83-84, pp. 61-69, 2019.

[33] T. G. Rebanowako, A. R. Yadav, and R. Joshi, "Age-Invariant Facial Expression Classification Method Using Deep Learning," presented at the Proceedings of 6th International Conference on Recent Trends in Computing, Singapore, 2021.

[34] G. Guo, R. Guo, and X. Li, "Facial Expression Recognition Influenced by Human Aging," *IEEE Transactions on Affective Computing,* vol. 4, no. 3, pp. 291-298, 2013.

[35] R. Angulu, J. R. Tapamo, and A. O. Adewumi, "Age estimation via face images: a survey," *EURASIP Journal on Image and Video Processing,* vol. 2018, no. 1, p. 42, 2018

[36] R. Angulu, J. R. Tapamo, and A. O. Adewumi, "Age-Group Estimation Using Feature and Decision Level Fusion," *The Computer Journal,* vol. 62, no. 3, pp. 346-358, 2018.

[37] Z. Lou, F. Alnajar, J. M. Alvarez, N. Hu, and T. Gevers, "Expression-Invariant Age Estimation Using Structured Learning," *IEEE Transactions on Pattern Analysis and Machine Intelligence,* vol. 40, no. 2, pp. 365-375, 2018.

[38] S. Motiian, M. Piccirilli, D. A. Adjeroh, and G. Doretto, "Unified deep supervised domain adaptation and generalization," presented at the Proceedings of the IEEE international conference on computer vision, 2017.

[39] P. Punyani, R. Gupta, and A. Kumar, "Neural networks for facial age estimation: a survey on recent advances," *Artificial Intelligence Review,* vol. 53, no. 5, pp. 3299-3347, 2020.

[40] M. M. Sawant and K. M. Bhurchandi, "Age invariant face recognition: a survey on facial aging databases, techniques and effect of aging," *Artificial Intelligence Review,* vol. 52, no. 2, pp. 981-1008, 2019.

[41] A. S. Osman Ali, V. Sagayan, A. M. Saeed, H. Ameen, and A. Aziz, "Age-invariant face recognition system using combined shape and texture features," *IET Biometrics,* vol. 4, no. 2, pp. 98-115, 2015.

[42] K. Baruni, N. Mokoena, M. Veeraragoo, and R. Holder, "Age Invariant Face Recognition Methods: A Review," presented at the 2021 International Conference on Computational Science and Computational Intelligence (CSCI), 15-17 Dec. 2021, 2021.

[43] A. Juhong and C. Pintavirooj, "Face recognition based on facial landmark detection," presented at the 2017 10th Biomedical Engineering International Conference (BMEiCON), 31 Aug.-2 Sept. 2017, 2017.

[44] A. Chinnnaswamy, P. Kumar, and S. Aravind, "Age Group Estimation using Facial Features," *International Journal of Emerging Technologies in Computational and Applied Sciences,* 2014.

[45] A. Srivastava, "Estimation of Age Groups based on Facial Features," *International Journal of Engineering and Technical Research,* vol. 7, pp. 115-121, 2018.

[46] S. A. Rizwan, A. Jalal, and K. Kim, "An Accurate Facial Expression Detector using Multi-Landmarks Selection and Local Transform Features," presented at the 2020 3rd International Conference on Advancements in Computational Sciences (ICACS), 17-19 Feb. 2020, 2020.

[47] M. Murtaza, M. Sharif, M. AbdullahYasmin, and T. Ahmad, "Facial expression detection using Six Facial Expressions Hexagon (SFEH) model," presented at the 2019 IEEE 9th Annual Computing and Communication Workshop and Conference (CCWC), 7-9 Jan. 2019, 2019.

[48] A. Barman and P. Dutta, "Influence of shape and texture features on facial expression recognition," *IET Image Processing,* vol. 13, no. 8, pp. 1349-1363, 2019.

[49] K. X. Beh and K. M. Goh, "Micro-Expression Spotting Using Facial Landmarks," presented at the 2019 IEEE 15th International Colloquium on Signal Processing & Its Applications (CSPA), 8-9 March 2019, 2019.

[50] D. Gong, Z. Li, D. Lin, J. Liu, and X. Tang, "Hidden Factor Analysis for Age Invariant Face Recognition," presented at the 2013 IEEE International Conference on Computer Vision, 1-8 Dec. 2013, 2013.

[51] H. Li, H. Zou, and H. Hu, "Modified Hidden Factor Analysis for Cross-Age Face Recognition," *IEEE Signal Processing Letters,* vol. 24, no. 4, pp. 465-469, 2017.

[52] H. Bar and M. T. Wells, "On Graphical Models and Convex Geometry," *arXiv preprint arXiv:2106.14255,* 2021.

[53] D. Kollias, "ABAW: Valence-Arousal Estimation, Expression Recognition, Action Unit Detection & Multi-Task Learning Challenges," presented at the 2022 IEEE/CVF Conference on Computer Vision and Pattern Recognition Workshops (CVPRW), 2022. [Online].

[54] D. Kollias, "ABAW: Learning from Synthetic Data & Multi-task Learning Challenges," Cham, 2023.

[55] D. Kollias, A. Schulc, E. Hajiyev, and S. Zafeiriou, "Analysing Affective Behavior in the First ABAW 2020 Competition," presented at the 2020 15th IEEE International Conference on Automatic Face and Gesture Recognition, 16-20 Nov. 2020, 2020.

[56] D. Kollias, V. Sharmanska, and S. Zafeiriou, "Distribution Matching for Heterogeneous Multi-Task Learning: a Large-scale Face Study," *ArXiv,* vol. abs/2105.03790, 2021.

[57] D. Kollias and S. Zafeiriou, "Expression, Affect, Action Unit Recognition: Aff-Wild2, Multi-Task Learning and ArcFace," *ArXiv,* vol. abs/1910.04855, 2019.

[58] D. Kollias and S. Zafeiriou, "Affect Analysis in-the-wild: Valence-Arousal, Expressions, Action Units and a Unified Framework," *ArXiv,* vol. abs/2103.15792, 2021.

[59] D. Kollias and S. Zafeiriou, "Analysing Affective Behavior in the second ABAW2 Competition," presented at the 2021 IEEE/CVF International Conference on Computer Vision Workshops (ICCVW), 11-17 Oct. 2021, 2021.

[60] C. Grossard *et al.*, "Teaching Facial Expression Production in Autism: The Serious Game JEMImE," *Creative Education,* vol. Vol.10No.11, p. 20, 2019.

[61] H. Kumazaki *et al.*, "Job interview training targeting nonverbal communication using an android robot for individuals with autism spectrum disorder," *Autism,* vol. 23, no. 6, pp. 1586-1595, 2019.

[62] W.-T. Tsai, I.-J. Lee, and C.-H. Chen, "Inclusion of third-person perspective in CAVE-like immersive 3D virtual reality role-playing games for social reciprocity training of children with an autism spectrum disorder," *Universal Access in the Information Society,* vol. 20, pp. 375-389, 2021.

[63] E. M. Medica, "Give me a kiss! An integrative rehabilitative training program with motor imagery and mirror therapy for recovery of facial palsy," *European journal of physical and rehabilitation medicine,* pp. 1-38, 2019.

[64] R. Okamoto, K. Adachi, and K. Mizukami, "[Effects of facial rehabilitation exercise on the mood, facial expressions, and facial muscle activities in patients with Parkinson's disease]," (in jpn), *Nihon Ronen Igakkai Zasshi,* vol. 56, no. 4, pp. 478-486, 2019.

[65] D. Kollias, V. Sharmanska, and S. Zafeiriou, "Face Behavior à la carte: Expressions, Affect and Action Units in a Single Network," *ArXiv,* vol. abs/1910.11111, 2019.

[66] D. Kollias *et al.*, "Deep Affect Prediction in-the-Wild: Aff-Wild Database and Challenge, Deep Architectures, and Beyond," *International Journal of Computer Vision,* vol. 127, pp. 907-929, 2018.

[67] S. Zafeiriou, D. Kollias, M. A. Nicolaou, A. Papaioannou, G. Zhao, and I. Kotsia, "Aff-wild: valence and arousal'In-the-Wild'challenge," presented at the Proceedings of the IEEE conference on computer vision and pattern recognition workshops, 2017.

[68] J. N. Kundu *et al.*, "Balancing discriminability and transferability for source-free domain adaptation," in *International Conference on Machine Learning*, 2022: PMLR, pp. 11710-11728.

[69] S. Pei, J. Sun, S. Xiang, and G. Meng, "Domain Decorrelation with Potential Energy Ranking," *arXiv preprint arXiv:2207.12194,* 2022.

[70] L. N. Smith, "Cyclical learning rates for training neural networks," presented at the 2017 IEEE winter conference on applications of computer vision (WACV), 2017.

[71] D. Kollias, "ABAW: learning from synthetic data & multi-task learning challenges," presented at the Computer Vision–ECCV 2022 Workshops: Tel Aviv, Israel, October 23–27, 2022, Proceedings, Part VI, 2023.

[72] S. M. Lundberg and S.-I. Lee, "A unified approach to interpreting model predictions," presented at the Proceedings of the 31st International Conference on Neural Information Processing Systems, Long Beach, California, USA, 2017.

[73] G. Erion, J. D. Janizek, P. Sturmfels, S. M. Lundberg, and S.-I. Lee, "Improving performance of deep learning models with axiomatic attribution priors and expected gradients," *Nature Machine Intelligence,* vol. 3, no. 7, pp. 620-631, 2021.

**Megan A. Witherow** received her B.S. degree in computer engineering from Old Dominion University (ODU), Norfolk, VA, USA in 2018. She is currently a PhD candidate at the Vision Laboratory, Dept. of Electrical and Computer Engineering, ODU, and a 2020 NSF Graduate Research Fellow. Her research interests include computer vision, deep learning, human-computer interaction, and affective computing.

**Manar D. Samad** received the M.S. degree in computer engineering from the University of Calgary, Calgary, AB, Canada in 2011, and the Ph.D. degree from Old Dominion University, Norfolk, VA, USA in 2016. He worked as a post-doctoral fellow at Geisinger Medical Center in Danville, PA for two years. He is currently an Assistant Professor in the Department of Computer Science at Tennessee State University, Nashville, TN, USA. His research interests include machine learning, health informatics, computer vision, and natural language processing.

**Norou Diawara** is Professor of Statistics in the Mathematics and Statistics Department at Old Dominion University, Norfolk, VA, USA. Prof. Diawara received his B.S. at the University Cheick Anta Diop in Dakar, Senegal; Maîtrise in Mathematics at University of Le Havre, France; Master's in Statistics at University South Alabama; and Ph.D. in Statistics at Auburn University, AL in 2006. His research areas are in estimation techniques of time to event data analyses and neighborhood level causal effects. Such research interests may be included in choice models, statistical pattern recognition using copulas and spatial-temporal models.

**Haim Y. Bar** is an associate professor in statistics at the University of Connecticut. He has a Ph.D. in statistics (Cornell), M.Sc. in computer science (Yale), and B.Sc. in mathematics (the Hebrew University in Jerusalem). His professional interests include statistical modeling, shrinkage estimation, high throughput applications, Bayesian statistics, variable selection, and machine learning. From 1995 to 1997, he was with Motorola, Israel, as a computer programmer. From 1997 until 2003 he worked for MicroPatent, LLC, where he held the position of Director of Software Development. In 2003, he moved to Ithaca, NY, and worked as a Principal Scientist at ATC-NY.

**Khan M. Iftekharuddin** (SM'02) received the B.Sc. degree in electrical and electronic engineering from the Bangladesh Institute of Technology, Dhaka, Bangladesh, in 1989, and the M.S. and Ph.D. degrees in electrical and computer engineering from the University of Dayton, Dayton, OH, USA, in 1991 and 1995 respectively. He is currently the Associate Dean for Research and Innovation with the Batten College of Engineering and Technology, Old Dominion University (ODU), Norfolk, VA, USA, and the Director of the ODU Vision Laboratory. His current research interests include computational modeling of intelligent systems and reinforcement learning, stochastic medical image analysis, intersection of bioinformatics and medical image analysis, distortion-invariant automatic target recognition, biologically inspired human and machine centric recognition, and machine learning for robotics.